\pdfoutput=1

\documentclass[11pt]{article}

\usepackage{EMNLP2022}

\usepackage{times}
\usepackage{latexsym}

\usepackage[T1]{fontenc}

\usepackage[utf8]{inputenc}

\usepackage{microtype}

\usepackage{graphicx}
\usepackage{booktabs}
\usepackage{multirow}
\usepackage{xltabular}
\usepackage{enumitem}
\usepackage{paralist}
\usepackage{hyperref}

\title{\textsc{Phee}: A Dataset for Pharmacovigilance Event Extraction from Text}

\author{
    Zhaoyue Sun\textsuperscript{\rm1}, 
    Jiazheng Li\textsuperscript{\rm1}, 
    Gabriele Pergola\textsuperscript{\rm1},
    Byron C. Wallace\textsuperscript{\rm2} \\
    \textbf{Bino John}\textsuperscript{\rm3}\textbf{,}
    \textbf{Nigel Greene}\textsuperscript{\rm3}\textbf{,}
    \textbf{Joseph Kim}\textsuperscript{\rm3} \and
    \textbf{Yulan He}\textsuperscript{\rm1,4,5}\\
  \textsuperscript{1}Department of Computer Science, University of Warwick \\
  \textsuperscript{2}Khoury College of Computer Sciences, Northeastern University \\
  \textsuperscript{3}AstraZeneca\\
  \textsuperscript{4}Department of Informatics, King's College London\\
  \textsuperscript{5}The Alan Turing Institute\\
  \texttt{\{Zhaoyue.Sun, Jiazheng.Li, Gabriele.Pergola.1\}@warwick.ac.uk} \\
  \texttt{\{bino.john, nigel.greene, joseph.kim1\}@astrazeneca.com} \\
  \texttt{b.wallace@northeastern.edu, yulan.he@kcl.ac.uk} \\
  }

\begin{document}
\maketitle

\begin{abstract}
The primary goal of drug safety researchers and regulators is to promptly identify adverse drug reactions. Doing so may in turn prevent or reduce the harm to patients and ultimately improve public health. Evaluating and monitoring drug safety (i.e., \textit{pharmacovigilance}) involves analyzing an ever growing collection of spontaneous reports from health professionals, physicians, and pharmacists, and information voluntarily submitted by patients. In this scenario, facilitating analysis of such reports via automation has the potential to rapidly identify safety signals. Unfortunately, public resources for developing natural language models for this task are scant. We present \textsc{Phee}, a novel dataset for pharmacovigilance comprising over 5000 annotated events from medical case reports and biomedical literature, making it the largest such public dataset to date. We describe the hierarchical event schema designed to provide coarse and fine-grained information about patients' demographics, treatments and (side) effects. Along with the discussion of the dataset, we present a thorough experimental evaluation of current state-of-the-art approaches for biomedical event extraction, point out their limitations, and highlight open challenges to foster future research in this area\footnote{Our data and code is available at \href{https://github.com/ZhaoyueSun/PHEE}{https://github.com/ZhaoyueSun/PHEE}}. 
\end{abstract}

\section{Introduction}

Pharmacovigilance is the pharmaceutical science that entails monitoring and evaluating the safety and efficiency of medicine use, which is vital for improving public health \citep{WHO:04}. 
Unexpected adverse drug effects (ADEs) could lead to considerable morbidity and mortality \citep{lazarou:98}. It has been reported that more than half of ADEs are preventable \citep{gurwitz:00}. Pharmacovigilance is therefore important for detecting and understanding ADE-related events, as it may inform clinical practice and ultimately mitigate preventable hazards. 

Collecting and maintaining the clinical evidence for pharmacovigilance can be difficult because it requires time-consuming manual curation to capture emerging data about drugs \citep{thompson:18}. Much of this information can be found in unstructured textual data including medical literature, notes in electronic health records (EHR), and social media posts. 
Using NLP methods to discover and extract adverse drug events from unstructured text may permit efficient monitoring of such sources
\citep{nikfarjam:15,huynh:16,ju:20,wei:20}.

\begin{figure*}[htp] 
\centering
\includegraphics[width=1.91\columnwidth]{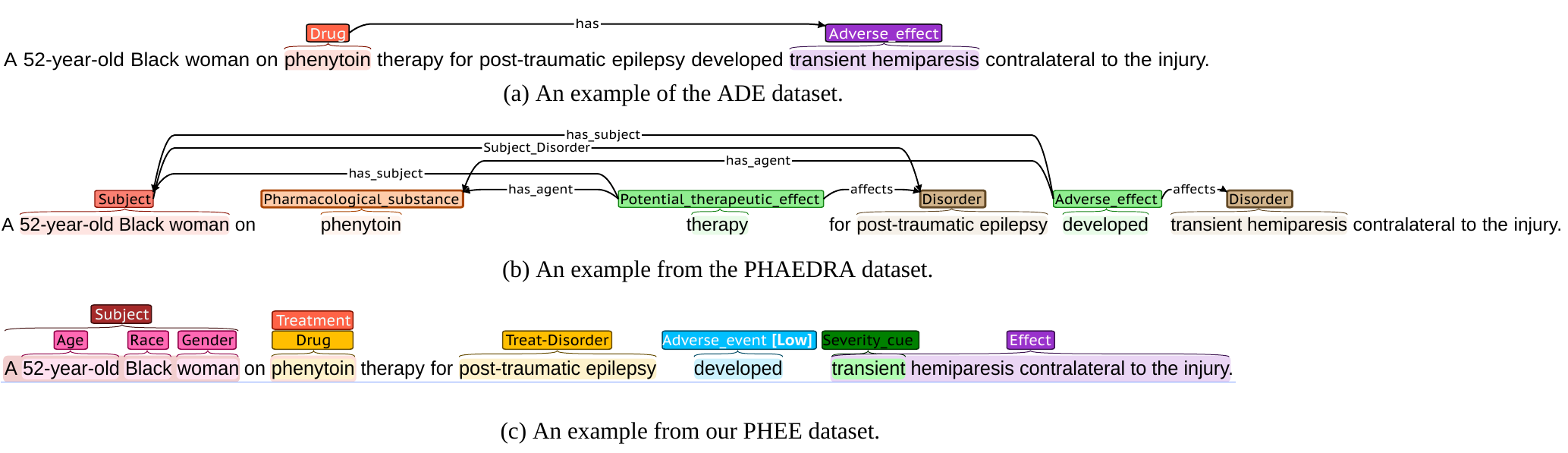} 
\caption{Comparison of annotations from (a) the ADE corpus, (b) the PHEADRA corpus and (c) our developed \textsc{Phee} corpus.} 
\label{fig:example} 
\end{figure*}

Past work has introduced pharmacovigilance corpora to support training and evaluation of NLP approaches for ADE extraction. However, most of these datasets (e.g., the ADE corpus; \citealt{gurulingappa:12}) contain annotations only on entities (such as drugs and side effects) and their binary relations as shown in Figure \ref{fig:example}(a). This ignores contextual information relating to human subjects, treatments administered, 
and more complex situations such as multi-drug concomitant use.  To address this problem, \citet{thompson:18} developed the PHAEDRA corpus, which includes annotations of not only drugs and side effects, but also subjects (human, specific species, bacteria, and so on) and events encoding descriptions of drug effects, which involve multiple arguments, 
and event attributes --- see Figure \ref{fig:example}(b). 

Despite these refinements, however, PHAEDRA does not provide detailed, nested annotations such as dosages, conditions, and patient demographic details. This granular information may provide critical context to clinical studies.
Furthermore, PHAEDRA consists of only 600 annotated abstracts of medical case reports
, making it challenging to train NLP models for pharmacovigilance events extraction since its annotations are in the document level and the actual annotated events are sparse. 

In this work we introduce a new annotated corpus, \textsc{Phee}, for adverse and potential therapeutic effect event extraction for pharmacovigilance study. 
The dataset consists of nearly 5,000 sentences extracted from MEDLINE case reports, and each sentence features two levels of annotations. 
With respect to coarse-grained annotations, each sentence is annotated with the event trigger word/phrase, event type and text spans indicating the event's associated subject, treatment, and effect. 
In a fine-grained annotation pass, further details are marked, such as patient demographic information, the context information about the treatments including drug dosage levels, administration routes, frequency, and attributes relating to events. An example annotation is shown in Figure \ref{fig:example}(c).

Using \textsc{Phee} as the benchmark, we conduct thorough experiments to assess the state-of-the-art NLP technologies for the pharmacovigilance-related event extraction task. We use sequence labelling and (both extractive and generative) QA-based methods as baselines and evaluate event trigger extraction and argument extraction. 
The extractive QA method performs best for trigger extraction with the exact match F1 score of 70.09\%, while the generative QA method achieves the best exact match F1 score of 68.60\% and 76.16\% for the main argument and sub-argument extraction, respectively.
Further analysis shows that current models perform well on average cases but often fail on more complex examples.

\textbf{Our contributions can be summarised as follows: }
\begin{inparaenum}[1)]
    \item We introduce \textsc{Phee}, a new pharmacovigilance dataset containing over 5,000 finely annotated events from public medical case reports. To the best of our knowledge, this is the largest and most comprehensively annotated dataset of this type to date.
    \item We collect hierarchical annotations to provide granular information about patients and conditions in addition to coarse-grained event information.
    \item We conduct thorough experiments to compare current state-of-the-art approaches for biomedical event extraction, demonstrating the strength and weaknesses of current technologies and use this to highlight challenges for future research in this area.
\end{inparaenum}


\section{Related Work}
\paragraph{Pharmacovigilance Related Corpora}
Prior pharmacovigilance-related corpora mainly has focused on annotation of entities (e.g., drugs, diseases, medications) and binary relations between them, namely, drug-ADE relations \citep{gurulingappa:12extraction,patki:14,ginn:14}, disorder-treatment relations \citep{rosario:04,roberts:09,uzuner:11,van:12}, and drug-drug interactions \citep{segura:11,boyce:12, rubrichi:12,herrero:13}. More recent open challenges, including the 2018 n2c2 shared task \citep{henry:20} and MADE1.0 challenge \citep{jagannatha:19}, have considered annotating additional relation types, such as drug-attribute and drug-reason relations, but they are still binary relationships. 

\citet{thompson:18} introduced the PHAEDRA corpus, extending the drug-ADE annotations to pharmacovigilance events. Compared to corpora that only annotate simple drug-ADE relations---referred to as \emph{AE} events in PHAEDRA---they further annotate three additional relations, namely the \emph{Potential Therapeutic Effect} (PTE) event which refers to the potential beneficial effects of drugs, 
the \emph{Combination} and the \emph{Drug-Drug Interaction} event which indicates multiple drug use and interactions between administered drugs, respectively.
In addition, PHAEDRA includes the subject as a type of named entities (NEs) and annotates three types event attributes, i.e., negated, speculated and manner. 
However, some key informative details are still missing in PHAEDRA. As the NE annotation of PHAEDRA is usually a single noun or a short noun phrase, detailed information about the subject (such as age and gender), and of the medication (e.g., dosage and frequency) is not captured. 

We set out to annotate a larger corpus with more detailed information to facilitate training of pharmacovigilance event extraction models. 
We build on existing corpora (PHAEDRA and ADE). 
The ADE corpus comprises $\sim$3,000 MEDLINE case reports and annotations on $\sim$4,000 sentences indicating adverse effects, but their annotations only involve drugs, dosages and adverse effects, and lack sufficient event details of interest. 
The PHAEDRA corpus reuses 227 abstracts from ADE and integrates an additional 370 abstracts (from other corpora and some novel entries). However, the PHAEDRA corpus is annotated at the document level, the actual annotated events are very sparse. We collected sentences in ADE and those in PHAEDRA with AE or PTE event annotations and enriched these using our proposed annotation scheme.



\paragraph{Biomedical Event Extraction}
Most existing biomedical event extraction methods work as ``pipelines'', treating trigger extraction and argument extraction as two stages \citep{bjorne:18, li:18, li:20, huang:20, zhu:20}; this can lead to error propagation. 
\citet{trieu:20} propose an end-to-end model that jointly extracts the trigger/entity and assigns argument roles to mitigate the problem of error propagation, but in contrast to our span-based annotation, this requires full annotation of all entities. 
\citet{ramponi:20} consider biomedical event extraction as a sequence labelling task, allowing them jointly model event trigger and argument extraction via multi-task learning. 

In other domains, recent work has formulated event extraction as a \emph{question answering} task \citep{du:20, li:20event, liu:20}. 
This new paradigm transforms the extraction of event trigger and arguments into multiple rounds of questioning, obtaining an answer about a trigger or an argument in each round. Such methods can reduce the reliance on the entity information for argument extraction and have proved to be data efficient. The current QA-based event extraction methods are mainly built on extractive QA which obtains the answer to a question by predicting the position of the target span in the original text. As such, a separate question needs to be formulated for different event and argument type. We also experiment with a generative QA method, which generates the answers directly, for comparison.

\section{The PHEE Dataset}
\subsection{Task Definition and Schema}
The PHEE corpus comprises sentences from biomedical literature annotated with information relevant to pharmacovigilance. 
Annotations are hierarchically structured in terms of textual \textit{events}. Following prior work \cite{thompson:18}, we define two main clinical event types: \emph{Adverse Drug Effect (ADE)} and \emph{Potential Therapeutic Effect (PTE)}, denoting potentially harmful and beneficial effects of medical therapies, respectively. 
%
Events consist of a \textit{trigger} and several \textit{arguments}, as defined by the ACE Semantic Structure \citep{LDC:05}. The trigger is a word or phrase that best indicates the occurrence of an event (e.g., `induced', `developed'), while the arguments specify the information characterizing an event, such as patient's demographic information, treatments, and \mbox{(side-)}effects (Figure \ref{fig:example}(c)).
We further organise arguments into two hierarchical levels, namely main and sub-arguments. Main arguments are longer text spans that contain the full description of an event aspect (e.g., \emph{treatment}),
while sub-arguments are usually words or short phrases included in main argument spans and highlighting specific details of the argument (e.g., \emph{drug}, \emph{dosage}, \emph{duration}, etc).

More specifically, in \textsc{Phee}, event arguments are defined as:
\begin{description}
  \item[Subject] highlights the patients involved in the medical event, with sub-arguments including \emph{age, gender, race,} \emph{number of patients} (labeled as \emph{population}) and \emph{preexisting conditions} (labeled as \emph{subject.disorder}) of the subject.
  \item[Treatment] describes the therapy administered to the patients, with sub-arguments specifying \emph{drug} (and their combinations), \emph{dosage}, \emph{frequency, route, time elapsed}, \emph{duration} and the \emph{target disorder}  (labeled as \emph{treatment.disorder}) of the treatment.
  \item[Effect] indicates the outcome of the treatment.
\end{description}

\noindent

We also collected annotations indicating three types of \emph{attributes} characterizing whether an event is \emph{negated}, \emph{speculated} or its \emph{severity} is indicated. See more details about the schema in Appendix \ref{apd:schema}.

\subsection{Data Collection and Validation}

\paragraph{Data Collection} To compose the \textsc{Phee} corpus, we collect existing medical case report abstracts from the ADE \citep{gurulingappa:12} and PHAEDRA \citep{thompson:18} datasets. 
We extract sentences from the abstracts and annotate them containing at least one adverse or therapeutic effect (\textit{ADE} or \textit{PTE}) event, for a total of over 4.8k sentences after deduplication.



\paragraph{Annotation Process} 
We hired 15 annotators in total to participate in our annotation, who are PhD students in the computer science or medical domain. We consulted our annotation schema with pharmacovigilance researchers and biomedical NLP researchers before starting the annotation. 

We conducted the corpus annotation through two stages to reduce the difficulty in dealing with medical text. In the first stage, we provided the annotators with sets of single sentences and asked them to highlight the event triggers and the text spans functioning as main arguments (i.e., \textit{subject}, \textit{treatment} and \textit{effect}). Each annotator annotates about 330 sentences during this stage. In the next stage, we randomly assigned the annotated sentences to different annotators who were required to verify the correctness of the previous annotations. Once confirmed, the annotations were expanded specifying the possible sub-arguments (e.g., for \textit{subjects}: \textit{age}, \textit{gender}, \textit{population}, \textit{race}, \textit{subject.disorder}), and attributes (e.g., \textit{negation}).
To ease the cognitive demand required to highlight fine-grained sub-arguments during the second stage, the annotators were split into three groups, each specialising in just one of the three main argument types. Specifically, four annotators are allocated for subject sub-argument annotation and four for effect and attribute annotation, while seven annotators are allocated for treatment sub-argument annotation due to the task complexity. Each annotator is responsible for around 1.4k or 700 instances during this stage. Additional notes on the annotation process can be found in the Appendix \ref{apd:ann-supplement}.



\paragraph{Data Validation} To ensure quality annotations, 
each stage of annotation was proceeded by several rounds of annotation trials, after which we discussed frequent inconsistencies.
When questions about specific instances surfaced during the annotation process, annotators flagged these sentences for review. 
While the main annotations of stage one were double-checked by the annotators in stage two, we randomly duplicated 20\% of the stage-two samples and assigned them to different groups to measure Inter-Annotator Agreement (IAA). 

We compute F1-score\footnote{Traditional Cohen's Kappa as IAA evaluation is not applicable for span-level computation due to an unknown number of negative cases.
We therefore follow previous work \cite{thompson:18, gurulingappa:12} choosing the F1 score as the more relevant IAA measurement.} as a measure of agreement between annotators. We calculate F1 scores between the sets of duplicated cases by (arbitrarily) selecting one annotation set as a ``reference'' to the other.
Specifically, we adopted the  EM\_F1 (at span-level) and Token\_F1 (at token-level) metric which are explained in details in Section \ref{par:arg-eval}. We report agreement scores in Table \ref{tab:iaa}. 

Consistency across trigger and argument types is over 80\%, indicating the effectiveness of  two-stage approaches. 
Agreement on sub-arguments is lower, 
which is expected due to the higher complexity of fine-grained medical annotations.
In particular, we notice a difficulty in consistency over the annotation of \emph{duration} and \emph{time\_elapsed}. One type of common inconsistent cases is "generalized expressions" (e.g., "chronic", "long-term", "shortly after"), which are annotated by some annotators but ignored by others. In addition, it is easy for annotators to confuse these two types of annotation. For example, the phrase "48 months" in "48 months postchemotherapy" is mistakenly annotated to be \emph{duration}, which, however, is generally believed should be \emph{time\_elapsed}. Other less inconsistent sub-argument types including \emph{frequency} and \emph{subject.disorder}. For \emph{frequency}, inconsistent cases including generalized expressions (e.g., "repeated", "continuous") and certain specific expressions such as "0.32mg/kg/day" that some annotators prefer to annotate "0.32mg/kg" as \emph{dosage} and "/day" as \emph{frequency} while others prefer to annotate the whole span as \emph{dosage}. For \emph{subject.disorder}, conflicts exist in "neutral"  expressions that describe the subject's health condition but not necessarily to be a disorder, such as "pregnant" and "nondiabetic". Apart from the difficult cases, inconsistency also occurs in the choices of span boundaries, especially for long arguments, or sometimes due to accident mistakes.  
Attribute annotations are inconsistent, probably due to their rarity in the corpus. 
\begin{table}
\centering
\resizebox{.7\columnwidth}{!}{
\begin{tabular}{@{}lcc@{}}
\toprule
 & EM\_F1 & Token\_F1 \\ \midrule
Trigger & 88.17 & 88.81 \\
Main-argument & 84.57 & 90.14 \\
Sub-argument & 80.25 & 83.24 \\
Attribute & 46.41 & 48.04 \\ \bottomrule
\end{tabular}}
\caption{Inter-Annotator Agreement (IAA).}
\label{tab:iaa}
\end{table}

\subsection{Dataset Statistics and Analysis}

\textsc{Phee} includes a total of 4,827 sentences and 5,019 annotated events.  
This makes \textsc{Phee} the largest annotated dataset on adverse drug events of which we are aware. 
We randomly divided train, dev, and test splits based on documents. 
Details about these splits are provided in Table \ref{tab:corpus-stat}.
\begin{table}
\centering
\resizebox{.7\columnwidth}{!}{
\begin{tabular}{lrrrr}\toprule
                         & Train & Dev   & Test  & Total \\ \midrule
\# Sentences                & 2,898 & 961   & 968   & 4,827 \\
\# Events                   & 3,006 & 1,003 & 1,010 & 5,019 \\
\hspace{3mm}- ADE  & 2,710 & 886   & 889   & 4,485 \\
\hspace{3mm}- PTE & 296   & 117   & 121   & 534  \\\bottomrule
\end{tabular}}
\caption{Dataset statistics on train/dev/test sets.}
\label{tab:corpus-stat}
\end{table}

Table \ref{tab:main-args} reports  statistics of the main event arguments. 
In general, each event contains at most one main argument of a particular type, but arguments might be discontinuous, leading to multiple spans representing a single argument. 
The average number of tokens per argument is about 3-4, which is generally longer than other datasets focusing only on biomedical entities (drugs, diseases or effects). 
\begin{table}
\centering
\resizebox{1\columnwidth}{!}{
\begin{tabular}{l|cccc}\toprule
          & \# ann. & \# spans & \# ann./sentence & avg. tokens/ann.  \\ \midrule
Subject   & 2,424        & 2,502 & 0.50 & 3.95             \\
Treatment & 5,018        & 5,329 & 1.04 & 3.25             \\
Effect    & 4,593        & 4,871 & 0.95 & 3.67             \\\bottomrule     
\end{tabular}}
\caption{Main argument statistics.}
\label{tab:main-args}
\end{table}

\begin{figure}
\centering
\includegraphics[scale=0.48]{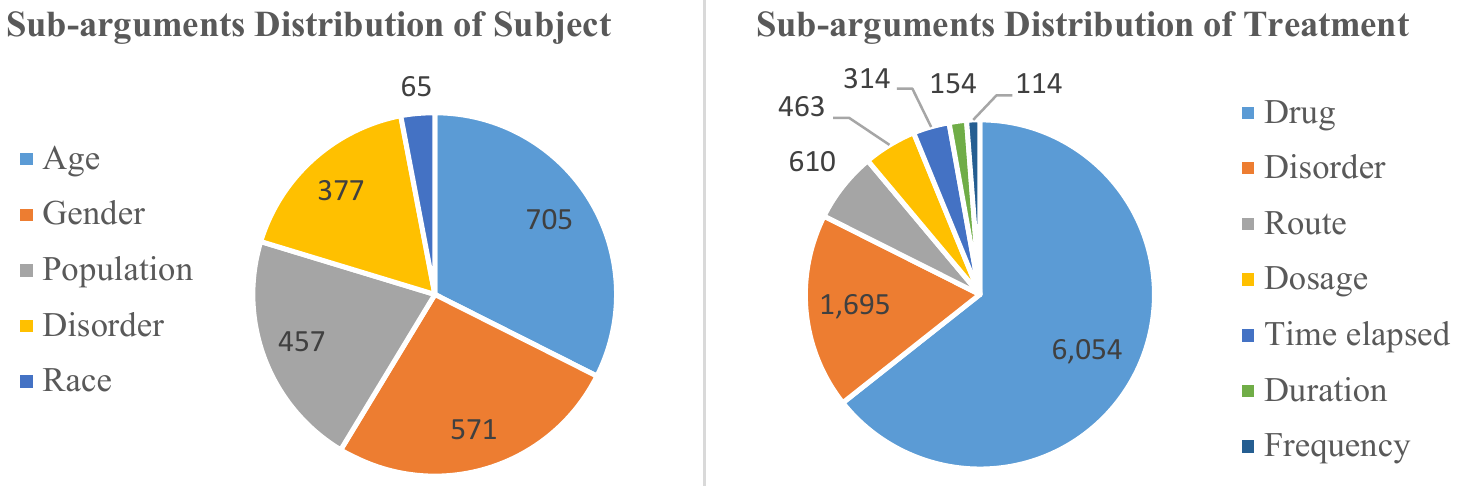} 
\caption{Distribution of sub-arguments.}
\label{fig:subarg-stat}
\end{figure}

Statistics about sub-argument annotations are provided in Figure \ref{fig:subarg-stat}. 
For the sub-arguments of the \textit{subject}, \textit{age} is the most frequently mentioned feature.  \textit{Gender}, \textit{population} and \textit{subject.disorder} are also comparatively common; \emph{race} is the rarest attribute. 
For \textit{treatment}, \textit{drug} names are the most frequently mentioned, even higher than the number of \textit{treatment} arguments due to the administered combinations of drugs.
The target \textit{disorder} of the treatment is the second most mentioned, providing context information in which the therapeutic or adverse events occurred.
In contrast, the other treatment's sub-arguments occur less frequently, resulting in a rather imbalance argument distribution.

Statistics of attributes are in Appendix \ref{apd:attr-stat}.

\begin{figure*}
\centering
\includegraphics[width=1.67\columnwidth]{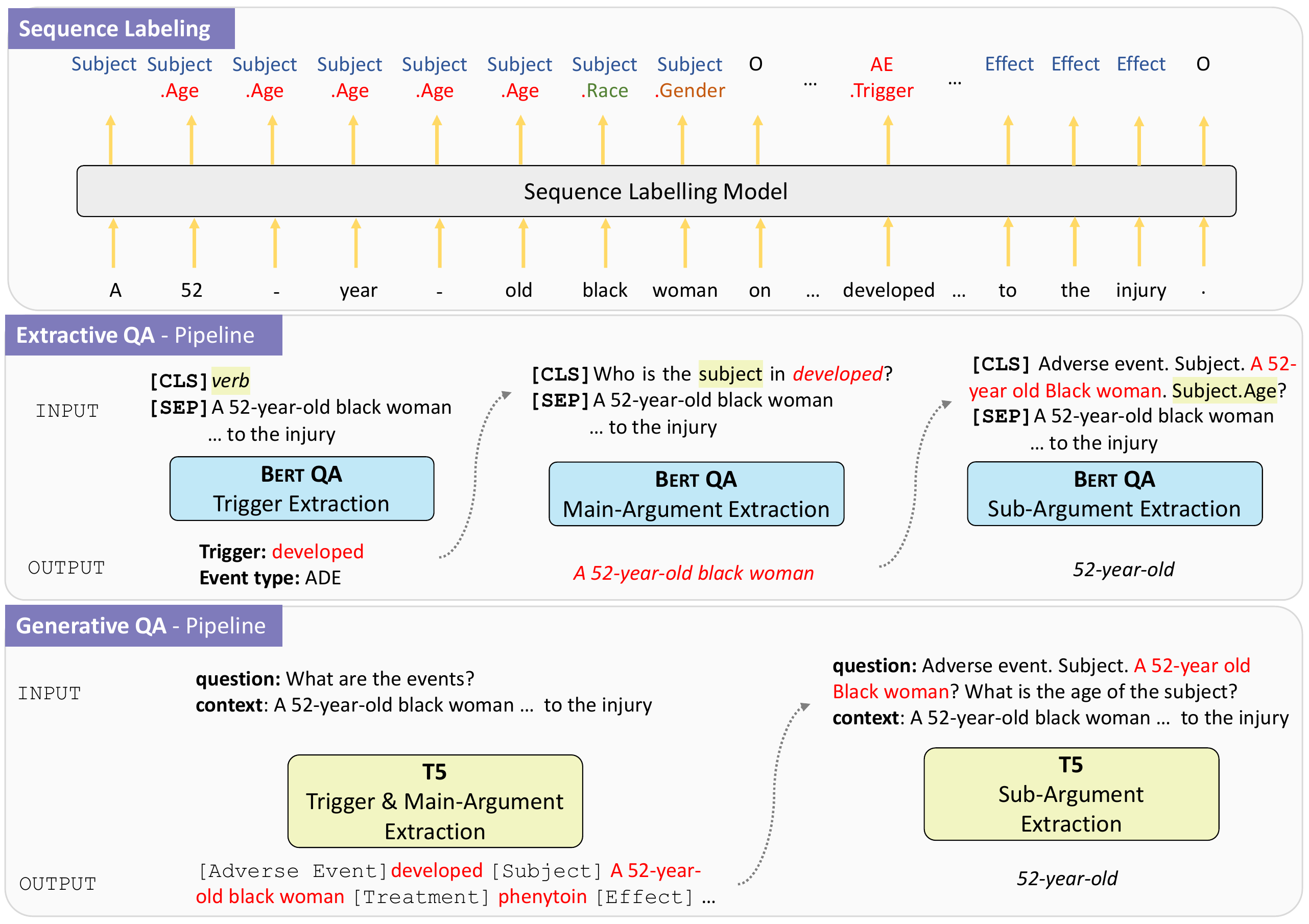} 
\caption{Illustrations of three baseline methods with the example: ``\textit{A 52-year-old Black woman on phenytoin therapy for post-traumatic epilepsy developed transient hemiparesis contralateral to the injury.}'' The diagram shows the extraction of the \textit{subject} for main argument extraction and the \textit{age} of the subject for sub-argument extraction.
\vspace{-10pt}} 
\label{fig:baseline} 
\end{figure*}

\section{Experiments}

We conduct evaluation of sequence labelling and QA-based methods (both extractive and generative) on our \textsc{Phee} dataset.
We describe our experimental design, evaluation metrics, and main results in this section. Reproduction details are in Appendix \ref{apd:reprod}.

\subsection{Models}

\paragraph{Sequence Labelling}
Given a sentence $S_i = \{x_1, x_2, ...x_j,..., x_n\}$, we encode event structures using token-level labels $Y_i = \{y_1, y_2, ...y_j,..., y_n\}$. 
We use the ``I-O'' scheme, in which the label ``I-X'' indicates a token is within a span of argument type $X$,  and ``O'' indicates it is outside of any argument span. 
As the main arguments and their associated sub-arguments usually overlap, we set the label to be ``I-A.B'' if the token is in a main argument span of type $A$ and a sub-argument span of type $B$ simultaneously. 
Correspondingly, the label will be ``I-A'' or ``I-B'' if the token only appears in a main argument or a sub-argument. 
For triggers, labels denote event types. An example of the flattened label sequence is shown in Figure \ref{fig:baseline}(a).

We use the ACE \citep{wang:20} model, which reaches state-of-the-art results for Named Entity Recognition (NER), as a representative sequence labelling method in our experiments. 

\paragraph{Extractive QA}
\label{sec:extract-qa}
We build our extractive QA model upon the EEQA method \citep{du:20}. 
Event triggers, main arguments, and sub-arguments are extracted in three sequential steps as shown in Figure \ref{fig:baseline}(b). 
We fine-tune the pre-trained BioBERT \citep{lee:20} base model on our dataset. 
In particular, the input is a sentence paired with a question, formatted as: 
`\texttt{[CLS] <question> [SEP] <sentence>}', where \texttt{<question>} and \texttt{<sentence>} are placeholders for a question template and an input sentence, respectively. The output is the text span extracted from the input sentence as answers. 
We experiment with different question templates for event triggers, main arguments and sub-arguments. 

For event trigger extraction, the model predicts a probability distribution across all events types (including a non-event case) for each input token based on BioBERT representations. 
Argument extraction is done for each argument type, 
where the probabilities of being the start/end position of an argument span are predicted for each token by a classification layer added on top of the BioBERT encoder. All possible <start, end> pairs are then filtered by thresholds of scores of the [CLS] token (which indicates a non-event prediction) to retrieve the extracted arguments. We also filter out spans that overlap with other spans with better scores.
We train the QA models for main-argument extraction and sub-argument extraction separately.


\paragraph{Generative QA}
Under the generative QA setting, we split the event extraction task into two stages.
In the first, event triggers and main arguments are extracted simultaneously.
In the second, sub-arguments are extracted. 
We fine-tune SciFive (PMC; \citealt{phan:21}) model, a T5 model pre-trained on Pubmed. 
An example of the input/output in the QA pipeline is shown in Figure \ref{fig:baseline}(c).

For the first stage, the question is simply `\emph{What are the events?}'. Each sentence is paired with the question in the form of `\texttt{question: <question> context: <sentence>}', where \texttt{question:} and \texttt{context:} are the fixed prompts, and \texttt{<question>} and \texttt{<sentence>} are placeholders for a pre-defined question and an input sentence, respectively. 
The gold-standard answer is constructed using a template `\texttt{[<event type>] <trigger> [<main argument type>] <main argument content> ...}', where \texttt{<$\cdot$>} is a placeholder to be replaced by the relevant content. For each event, the trigger comes first, followed by the main arguments in the order of \emph{subject}, \emph{treatment} and \emph{effect}. Multiple events are flattened into a sequence. The QA model then generates answers from which we obtain the event type, event trigger, and main argument spans via pattern matching. For the second stage, we use the questions defined for sub-arguments in extractive QA. The model input and gold-standard answers are formulated in a similar way as the first stage.




\subsection{Evaluation Metrics}

We evaluate model  performance on event trigger extraction and argument extraction separately. 
Punctuation and articles are ignored during evaluation.

\paragraph{Event trigger extraction} Following \citet{lin:20}, we use the F1 metric for the evaluation of event trigger identification and event trigger classification. Specifically, \emph{trigger identification} (Trig-I) evaluates how well the trigger words match their corresponding references; \emph{trigger classification} (Trig-C) evaluates not only the mentioning words but also event types. As the event trigger words could be ambiguous even for humans, and the 
detection of the presence of an ADE or PTE event is argubly more important, we further compute the \emph{event classification} (Event-C) F1 score, which evaluates whether the event type of a trigger word matches its reference.

\paragraph{Argument extraction}
\label{par:arg-eval}
Argument evaluation is also conducted from both \emph{identification} and \emph{classification} perspectives. Specifically, an argument span is correctly \emph{identified} if its event type and offsets match a gold-standard span, and it is correctly \emph{classified} if the argument type also matches. Considering that argument spans could be long and the exact match (i.e., span-level) evaluation might be too strict, we additionally report 
token-level evaluation results. Specifically, \emph{EM\_F1} measures the percentage of the predicted spans that match the ground truth spans exactly and \emph{Token\_F1} measures the average token overlap between the predictions and references. As there might be multiple spans for each argument, we compute both metrics by micro-averaging. That is, we accumulate the number of matched spans (or tokens) across the corpus as the True Positive (TP) value, and compute the precision, recall and F1 accordingly.

\subsection{Results and Analysis}
\label{sec:results}
We compare three families of baselines on the \textsc{Phee} dataset. For the extractive QA and the generative QA approaches, we explored several question templates and report only the results of templates which perform best on the development set. A more extensive analysis on different template formats is discussed in Appendix \ref{apd:templates}. 

\subsubsection{Evaluation: Trigger Extraction} Table \ref{tab:trig-extract} reports the performance of the three baselines on trigger extraction.  Extractive QA achieves the best result on both trigger identification and classification. 
However, it is worth mentioning that due to the EEQA design, only the first token of the trigger could be used for training and evaluation. 
Nevertheless, the comparison is still relevant as the trigger has the only linguistic function of representing an event occurrence but a limited semantic content.
Instead, the generative QA model obtained the best comparative performance when classifying the event type(s) of the whole sentence, independently of the particular trigger extracted.

\begin{table}
\centering
\resizebox{.66\columnwidth}{!}{
\begin{tabular}{@{}lccc@{}}
\toprule
 & Trig-I & Trig-C & Event-C \\ \midrule
Seq Labelling & 67.98 & 67.98 & 90.69 \\
Extractive QA & \textbf{70.71} & \textbf{70.09} & 85.61 \\
Generative QA & 68.25 & 68.25 & \textbf{95.16} \\ \bottomrule
\end{tabular}}
\caption{Results for trigger extraction.}
\label{tab:trig-extract}
\end{table}

\subsubsection{Evaluation: Argument Extraction} We present the main argument and sub-argument extraction results in Table \ref{tab:arg-extract}. Generative QA achieves the best results in both main argument and sub-argument extraction. Extractive QA performs better than sequence labelling in main argument extraction, but worse in sub-argument extraction. We sampled and analysed the error cases of the three approaches, and present some of them in Table \ref{tab:apd-example}. 


\begin{table}
\centering
\resizebox{0.95\columnwidth}{!}{
\begin{tabular}{@{}lcccc@{}}
\toprule
  & \multicolumn{2}{c}{Main-arguments} & \multicolumn{2}{c}{Sub-arguments} \\ \cmidrule(l){2-3} \cmidrule(l){4-5} 
   & EM\_F1 & Token\_F1 & EM\_F1 & Token\_F1 \\ \midrule
   \multicolumn{5}{c}{\emph{Argument Identification}}\\\midrule
 Sequence Labelling & 59.85 & 73.98 & 70.75 & 73.12 \\
 Extractive QA & 65.87 & 77.00 & 66.71 & 69.97 \\
 Generative QA & \textbf{68.85} & \textbf{81.63} & \textbf{77.33} & \textbf{78.38} \\ \midrule
 \multicolumn{5}{c}{\emph{Argument Classification}}\\\midrule
 Sequence Labelling & 59.61 & 73.16 & 68.88 & 69.31 \\
 Extractive QA & 65.70 & 75.92 & 64.98 & 66.69 \\
 Generative QA & \textbf{68.60} & \textbf{80.04} & \textbf{76.16} & \textbf{76.10} \\ \bottomrule
\end{tabular}}
\caption{Results for arguments extraction.}
\label{tab:arg-extract}
\end{table}

\begin{table*}[htb]
\centering
\resizebox{0.63\textwidth}{!}{
\begin{tabular}{@{}lcccccc@{}}
\toprule
 & \multicolumn{2}{c}{Seq Labelling} & \multicolumn{2}{c}{Extractive QA} & \multicolumn{2}{c}{Generative QA} \\ \cmidrule(l){2-3} \cmidrule(l){4-5} \cmidrule(l){6-7}
 & EM\_F1 & Token\_F1 & EM\_F1 & Token\_F1 & EM\_F1 & Token\_F1 \\\midrule
Subject & 39.72 & 66.57 & 64.26 & 75.13 & \textbf{68.72} & \textbf{80.77} \\
- Age & 82.91 & 88.06 & 82.07 & 89.68 & \textbf{89.25} & \textbf{93.48} \\
- Disorder & \textbf{30.26} & \textbf{38.99} & 21.36 & 25.17 & 24.19 & 33.07 \\
- Gender & 82.61 & 83.03 & 84.05 & 84.50 & \textbf{88.73} & \textbf{89.13} \\
- Population & 66.28 & 63.55 & 65.79 & 64.89 & \textbf{75.90} & \textbf{80.37} \\
- Race & 75.00 & 77.78 & 85.71 & 80.00 & \textbf{93.33} & \textbf{87.50} \\\midrule
Treatment & 58.85 & 70.28 & 61.35 & 71.97 & \textbf{63.66} & \textbf{75.80} \\
- Drug & 78.63 & 79.48 & 74.63 & 75.07 & \textbf{85.28} & \textbf{85.28} \\
- Disorder & 58.93 & 63.30 & 54.19 & 62.62 & \textbf{65.68} & \textbf{70.77} \\
- Route & 63.06 & 67.32 & 63.40 & 69.87 & \textbf{72.48} & \textbf{77.13} \\
- Dosage & 50.52 & 56.95 & 57.95 & 62.47 & \textbf{63.10} & \textbf{71.36} \\
- Time elapsed & \textbf{45.21} & \textbf{60.57} & 39.68 & 54.49 & 38.02 & 58.00 \\
- Duration & 22.54 & 42.67 & 27.59 & 37.70 & \textbf{30.77} & \textbf{45.26} \\
- Frequency & 36.92 & 42.62 & \textbf{54.55} & \textbf{53.85} & 51.16 & 50.67 \\
- Combination.Drug & 60.71 & 59.74 & 34.46 & 39.55 & \textbf{69.11} & \textbf{67.62} \\\midrule
Effect & 70.75 & 79.72 & 71.21 & 80.26 & \textbf{74.00} & \textbf{83.63} \\ \bottomrule
\end{tabular}}
\caption{Classification results for each argument type. Best results for each argument type are highlighted in bold.}
\label{tab:rst-type}
\end{table*}

In particular, for main argument extraction, we observe that a common error of extractive QA is the failure of detecting an event trigger in an early stage, making it skip extracting main arguments in the subsequent stages. Generative QA performs better probably because it extracts the trigger and main arguments simultaneously thus avoiding the problem of error propagation. 
For sequence labelling, the most prominent problem is the incompleteness of the extracted main arguments, especially for the \textit{subject} argument. 
One possible reason is that the main argument and sub-argument labels are flattened into one sequence, which results in the loss of the information about the relations between the main argument and its sub-arguments, therefore hurting the extraction performance. 

For the sub-argument extraction, the performance of the extractive QA drops to the worst, probably due to further error propagation from the previous two stages. For the other two methods, the sequence labelling method seems to be more severely affected by trigger extraction errors. In some cases, the argument spans are matched, but no trigger is detected in the sentence, thus leading to a failure. 
The generative QA method's performance at this stage is relatively less influenced by the main argument extraction results compared to the other two approaches, but we notice it could easily fail to extract less frequent sub-arguments.
One possible downside of generative QA models when used for information extraction is that they may 
generate tokens not in the original input sentence, 
but in our sampled cases, such errors are very rare.

\subsubsection{Evaluation for Each Argument Type} 

In Table \ref{tab:rst-type}, we present the results for each argument type. Firstly, among all main argument types, the \emph{effect} seems to be the easiest one to be extracted. This is probably due to its abundant occurrences and relatively distinct features compared to other argument types. Although the \emph{treatment} also occur frequently in the corpus, models perform much poorer on \emph{treatment} extraction. The main reason is that the length of the treatment spans varies, and the information of the treatment could be more complex, which leads to the fragmented extraction results. The \emph{subject} while appearing less frequently than \emph{treatment} and \emph{effect}, have relatively simpler linguistic patterns. As such, their exaction results are better than \emph{treatment} when using QA models. 

For the sub-arguments, highly frequent arguments with simpler linguistic patterns such as \emph{age, gender}, and \emph{drug} get promising results. Some arguments with relatively limited expressions, such as \emph{race} and \emph{frequency}, although very rare in our dataset, still have merit or moderate extraction result. Some sub-arguments such as \emph{subject.disorders} and \emph{treatment.disorders}, can be confusing even for human annotators, getting relatively low extraction performance. Models' performance on \emph{subject.disorder} is even poorer due to its less occurrence in the dataset. 
Another pair of arguments that are easily confused is \emph{time elapsed} and \emph{duration}, both of which contain temporal expressions. Combined with the low occurrence frequencies, these two arguments also get quite low extraction results.


\section{Challenges and Future Directions}
Our analysis of experimental results, suggests the following open challenges for the extraction of pharmacovigilance events. Firstly, the models perform poorly on arguments with similar entity mentions but different argument roles. For example, a disease mentioned in text could be annotated as \emph{treatment.disorder} if it is the target of the treatment or \emph{subject.disorder} if it refers to someone's disease but not targeted for treatment. A similar problem can be observed for arguments of  temporal expressions such as \emph{time elapsed} and \emph{frequency}. 
The poor performance on such arguments seems to indicate that existing models are not able to perform deep semantic analysis. Additional constraints encoding linguistic constructs between entity mentions and main argument types could be explored to guide the event extraction model through, for example, posterior regularisation. 

Secondly, the models' performance on argument types with limited annotated training instances deteriorates drastically. 
One path forward is therefore to explore efficient few-shot learning strategies to improve models' generalisability on rare argument types. Also, there might exist corpora annotated with similar argument roles but for different purposes, for example, the corpora for medication extraction \cite{jagannatha:19} where drug dosage and frequency are annotated. It is possible to leverage external drug or disease knowledge through knowledge distillation. 

Finally, none of the existing models cope well with the presence of multiple events in a sentence. This is mainly because existing annotations rely heavily on event triggers to differentiate events and require explicit linking between arguments and their respective event triggers. 
However, trigger identification itself is ambiguous and difficult even for human annotators. In some cases, multiple events could share the same trigger. 
For pipeline-based models, i.e., the QA models in our work, detection of multiple triggers is prone to error, thus making it hard for subsequent argument extraction due to error propagation. For the sequence labelling model, it is difficult to flatten the annotations of multiple events into a single label sequence. We thus duplicate the multi-event cases during training, and only provide a single event annotation for each case at one time. However, it becomes impossible to obtain full extract results for multiple events during the inference stage. In the future, rather than sequence labelling or QA-based extraction approaches, it is worth exploring graph-based approaches for multi-event extraction in which entity mentions are nodes in the graph while event extraction can be framed as soft clustering of entity mentions.


\section{Conclusion}
In this paper, we present the development of a novel corpus, \textsc{Phee}, composed of sentences from the medical case reports annotated with pharmacovigilance-related events. Events in \textsc{Phee} are hierarchically annotated with coarse and fine-grained information about patient demographics, treatments, and (side) effects. We use it to evaluate state-of-the-art NLP models for pharmacovigilance event extraction. Experimental results show that current models could capture reasonable information in common cases but face challenges for complex situations such as distinguishing semantically-similar arguments, dealing with the low resource setting, and extracting multi-events from text. 

\section*{Limitations}
This study has several limitations. First, despite the implementation of a quality control process, the collected annotations inevitably have some quality issues. For example, 
to reduce cognitive load, we split the annotation process into two stages and required  annotators working on the second-stage annotation to check and correct first-stage annotations. However, we noticed that many annotators are more willing to keep the previous annotations as they are unless the errors can be easily identified. This may lead to inflated IAA results. 

Also, although trained for the task, the lack of medical background of annotators may have some impact on the quality of the dataset. Second, our dataset only contains two event types, Adverse Drug Event (ADE) and Potential Therapeutic Effect (PTE). It is worth considering adding sentences with the \emph{null} event type, that is, not associated with ADE or PTE. Furthermore, only one base PLM model for each baseline was chosen in our experiments, and more encoding methods are worth exploring in the future. Finally, although we have provided the annotations of event attributes such as \emph{speculation}, \emph{negation} and \emph{severity}, we have not implemented baseline models for event attribution detection, partly due to the few annotated cases. In the future, we will explore semi-supervised learning approaches for event attribute detection.



\section*{Ethics}
We used abstracts of publicly published medical reports as data sources in which no patient sensitive information is revealed. 
It should be noted that the adverse events and potential therapeutic events presented in this work are only based on textual-level information extraction and do not necessarily indicate any causal relations between drugs and effects. Causality assessment for pharmacovigilance should follow a rigorous assessment framework such as the assessment criteria of various causality categories defined in the WHO-UMC system\footnote{\url{https://who-umc.org/media/164200/who-umc-causality-assessment_new-logo.pdf}}.

\section*{Acknowledgements}

This work was supported in part by the UK Engineering and Physical Sciences Research Council (grant no. EP/T017112/1, EP/V048597/1, EP/X019063/1), and the National Science Foundation (NSF) grant 1750978. YH is supported by a Turing AI Fellowship funded by the UK Research and Innovation (grant no. EP/V020579/1). 

\bibliography{anthology,custom}
\bibliographystyle{acl_natbib}

\newpage
\appendix

\setcounter{table}{0}
\renewcommand{\thetable}{A\arabic{table}}

\section*{Appendix}

\section{Annotation Schema}
\label{apd:schema}
We present a full list of the definitions of all items we annotated.
\paragraph{Event} An event is annotated with an event \emph{trigger}, several \emph{arguments} and \emph{attributes} (if any). We annotate the following types of events:
\begin{description}
  \item[\emph{Adverse Drug Effect:}] The use of a drug or combination of drugs cause a harmful effect on the human patient.
  \item[\emph{Potential Therapeutic Effect:}] The use of a drug or combination of drugs bring a potential beneficial effect on the human patient.
  \item[\emph{Combination (sub-event):}] More than one drug is treated for the patient. The combination sub-event consists of a trigger and several drug arguments. It usually plays the role of a sub-argument under the \emph{treatment} argument of an ADE/PTE.
\end{description}

\paragraph{Arguments} An argument describes the information characterizing an event.
\begin{description}
  \item[\emph{Subject:}] highlights the patients involved in the medical event. Sub-arguments of \emph{subject} are:
  \begin{description}
  \item[\emph{Subject.Age:}] concrete age or span that indicates an age range.
  \item[\emph{Subject.Gender:}] the span that indicates the subject’s gender.
  \item[\emph{Subject.Population:}] the number of patients receiving the treatment.
  \item[\emph{Subject.Race:}] the span that indicates the subject’s race/nationality.
  \item[\emph{Subject.Disorder:}] preexisting conditions, i.e., disorders that the subject suffers other than the target disorder of the treatment.
  \end{description}
  \item[\emph{Treatment:}] describes the therapy administered to the patients.
  \begin{description}
  \item[\emph{Treatment.Drug:}] drugs used as therapy in the event.
  \item[\emph{Treatment.Dosage:}] the amount of the drug is given.
  \item[\emph{Treatment.Frequency:}] the frequency of drug use.
  \item[\emph{Treatment.Route:}] the route of drug administration.
  \item[\emph{Treatment.Time\_elapsed:}] the time elapsed after the drug was administered to the occurrence of the (side) effect.
  \item[\emph{Treatment.Duration:}] how long the patient has been taking the medicine (usually for long-term medication).
  \item[\emph{Treatment.Disorder:}] the target disorder of the medicine administration.
  \end{description}
  \item[\emph{Effect:}] indicates the outcome of the treatment.
\end{description}

\paragraph{Attribute} \emph{Attributes} interpret certain properties of events, i.e., indicating whether an event is \emph{negated} or \emph{speculated}, and the \emph{severity} level of the event.  
\begin{description}
  \item[\emph{negated:}] the attribute \emph{negated} denotes whether or not there is any textual cues showing the event is negated, i.e., for ADE, the adverse effect does not exist; or for PTE, the therapy is ineffective.
  \item[\emph{speculated:}] the attribute \emph{speculated} indicates if there is any uncertain or speculation as to whether an event will actually happen. Considering the speculative nature of the medical case reports, we only annotate a \textit{speculated} attribute when the speculative attitude of the author is explicitly remarked. 
  \item[\emph{severity:}] the attribute \emph{severity} refers to the severity level of the adverse effect. For example, the fatal effect is a `high severity', while a minor symptom could be a `low severity'. In general, we do not annotate `severity' for PTE events.
\end{description}

\section{Annotation Process Supplement}
\label{apd:ann-supplement}
To facilitate annotation we used \emph{brat} \citep{stenetorp:12}, a web-based tool. 
Annotators were instructed to mark trigger and argument spans as \textit{brat-entities}, and discontinuous spans as \textit{brat-fragments}. As for the case where a sentence contains multiple events, each argument will be connected to the trigger of its corresponding event with \textit{brat-links}. 

All annotators are volunteering and paid by the hour. In total we hired 15 annotators that spent around 20 and 30 hours per person on the first and the second stage of annotation, respectively.

\section{Statistics of Attributes}
\label{apd:attr-stat}

The occurrence of attributes is relatively rare in our dataset, and their statistics are illustrated in Table \ref{tab:attribute-stat}. Specifically, although severity annotations, which refer to the severity level of the adverse effect, are helpful in adverse effect vigilance, it can be seen that there are very few mentions of this attribute in the dataset.

\begin{table}
\centering
\resizebox{.7\columnwidth}{!}{
\begin{tabular}{ccc}\toprule
\# Speculated & \# Negated & \# Severity \\ \midrule
633           & 412        & 76         \\ \bottomrule
\end{tabular}}
\caption{Distribution of event attributes.}
\label{tab:attribute-stat}
\end{table}

\section{Training Details and Hyperparameter Setting}
\label{apd:reprod}
\paragraph{Experiments $-$ Sequence Labelling}
For sequence labelling experiments, we use the code of ACE\footnote{\url{https://github.com/Alibaba-NLP/ACE}}\cite{wang:20}. ACE develops a neural architecture search algorithm to automatically find better concatenations of transformer-based embeddings. With limited computing resources, we choose BERT (Base Cased, 1.1M parameters; \citealt{kenton:19}) and BioBERT (Base v1.1, 1.1M parameters; \citealt{lee:20}) as the base embeddings. We firstly fine-tune the BERT and BioBERT separately on our data, and run the ACE algorithm with fixed fine-tuned embeddings. We follow the default released hyper-parameter setting of the ACE algorithm. When fine-tuning the embeddings, the AdamW \cite{loshchilov:18} optimizer is used with a learning rate of $5\times10^{-5}$ and the model is trained for $20$ epochs. For training the ACE controller, we use the Stochastic Gradient Descent (SGD) optimizer with an initial learning rate of $0.1$. The controller will anneal the learning rate by 0.5 if there is no improvement on the development set for 5 epochs. The batch sizes for both embedding fine-tuning and controller training are 32.
We run the experiments on an NVIDIA TITIAN RTX GPU. The PLM fine-tuning costs about 2 hours for each model, while the ACE controller training costs about one day for a maximum of 150 epochs run.

\paragraph{Experiments $-$ Extractive QA}
For extractive QA experiments, we fine-tune the EEQA\cite{du:20} model on our data\footnote{\url{https://github.com/xinyadu/eeqa}}. We use the BioBERT (Base Cased, 1.1M parameters; \citealt{kenton:19}) as the base model. The SGD algorithm is used as the optimizer, and the learning rate is set to $1\times10^{-5}$, $5\times10^{-5}$, $5\times10^{-5}$ for trigger extraction, main argument extraction and sub-argument extraction, respectively. We use a batch size of 32 for trigger extraction and 16 for argument extraction. We set the maximum training epochs to $10$. Experiments are conducted on an NVIDIA TITIAN RTX GPU. Training time for trigger extraction, main argument extraction and sub-argument extraction are about 0.5, 1, 4 hours, respectively. The training time varies according to the number of argument(or trigger) types that need to be asked for each instance.

\paragraph{Experiments $-$ Generative QA}
For generative QA experiments, we run the experiments with the Huggingface example code for question answering\footnote{\url{https://github.com/huggingface/transformers/tree/main/examples/pytorch/question-answering}}. We fine-tune the SciFive (PMC Base, 2.2M parameters; \citealt{phan:21}) model, a T5 model pre-trained on a large-scale Pubmed corpus, on our dataset. The training batch size is 16. The learning rate is $5\times10^{-4}$ for main argument extraction and $5\times10^{-5}$ for sub-argument extraction. We train the model for no more than 20 epochs with early stopping patience as 2 epochs. We use beam search for decoding with the beam size of 3. We use an NVIDIA TITIAN RTX GPU for model training. The training of the generative QA model costs one hour (or less) and about four hours for (trigger and) main arguments extraction and sub-arguments extraction, respectively.

\section{Experimental Results using Different Question Templates}
\label{apd:templates}
We present the experimental results on the development set with different question templates below. Table \ref{tab:temp-trigger} and Table \ref{tab:temp-main} show the results of the extractive QA model when using different templates for trigger extraction and main argument extraction, respectively. Table \ref{tab:temp-sub} shows the sub-argument extraction results of the extractive QA and the generative QA method. 

Overall, we observe that using different templates does not have a large impact on the results, which is probably due to the fact that our dataset involves few event types and relatively fixed argument types. 
Specifically, templates that achieve the best results on different metrics vary. The query template \textit{verb} obtains the best trigger and event type detection performance, while a full-sentence question \textit{What is the trigger in the event?
} get modestly better trigger identification performance.
For main argument extraction, the template including a query of the argument and the event trigger achieves the best scores.
For sub-argument extraction, the best-performing templates also slightly differ depending on the model. 
For the extractive QA model, a brief question including information on the event type, the main argument type and extracted span, and the queried sub-argument type gets the best exact match result, while also giving this information but changing the query argument type to a complete sentence would achieve the best token-level score. For the generative QA model, the argument type-specific query with all information about the event type and the main argument performs best on both span-level and token-level evaluation.


\begin{table*}[h]
\centering
\resizebox{.8\textwidth}{!}{
\begin{tabular}{@{}lccc@{}}
\toprule
Template & Trigger\_CLS\_F1 & Trigger\_IDT\_F1 & Event\_CLS\_F1 \\ \midrule
What is the trigger in the event? & 74.14 & \textbf{75.34} & 87.24 \\
What happened in the event? & 73.76 & 74.97 & 87.64 \\
trigger & 73.67 & 74.87 & 87.14 \\
t & 73.72 & 74.73 & 87.59 \\
action & 72.98 & 74.48 & 87.24 \\
verb & \textbf{74.18} & 75.00 & \textbf{87.81} \\
null & 73.61 & 75.04 & 86.68 \\ \bottomrule
\end{tabular}
}
\caption{Trigger extraction results with different templates for the extractive QA method. }
\label{tab:temp-trigger}
\end{table*}


\begin{table*}
\centering
\begin{tabular}{@{}lcccc@{}}
\toprule
 \multirow{2}{*}{Template} & \multicolumn{2}{c}{Identification} & \multicolumn{2}{c}{Classification} \\ \cmidrule(l){2-3} \cmidrule(l){4-5}
 & EM\_F1 & Token\_F1 & EM\_F1 & Token\_F1 \\ \midrule
\textless argument type \textgreater  & 70.35 & 83.60 & 70.27 & 81.67 \\
\textless argument type \textgreater $\:$ in \textless{}event type\textgreater{} & 70.71
 & 82.83 & 70.60 & 81.67 \\
\textless{}argument type\textgreater$\:$in \textless{}event trigger\textgreater{} & 
71.80 & 84.11 & 71.64 & 82.68 \\
\textless{}argument query\textgreater{} & 69.82 & 82.83 & 69.58 & 81.32 \\
\textless{}argument query\textgreater$\:$in \textless{}event type\textgreater{} & 70.20 & 83.28 & 70.08 & 82.12 \\
\textless{}argument query\textgreater$\:$in \textless{}event trigger\textgreater{} & \textbf{72.33} & \textbf{85.06} & \textbf{72.17} & \textbf{83.57} \\ \bottomrule
\end{tabular}
\caption{Main argument extraction results with different templates for the extractive QA method. }
\label{tab:temp-main}
\end{table*}

\begin{table*}
\centering
\resizebox{.8\textwidth}{!}{
\begin{tabular}{@{}lcccc@{}}
\toprule
\multirow{2}{*}{Template} & \multicolumn{2}{l}{Extractive QA} & \multicolumn{2}{l}{Generative QA} \\
\cmidrule(l){2-3} \cmidrule(l){4-5} & EM\_F1 & Token\_F1 & EM\_F1 & Token\_F1 \\\midrule
\textless{}sub-argument type\textgreater{} & 71.21 & 74.26 & 71.94 & 79.15 \\\midrule
\textless{}sub-argument type\textgreater$\:$in \textless{}event type\textgreater{} & 72.53 & 76.81
 & 73.16 & 80.89 \\\midrule
\textless{}sub-argument type\textgreater$\:$in \textless{}main argument span\textgreater{} & 
76.14 & 77.56 & 77.13 & 84.19 \\\midrule
\begin{tabular}[c]{@{}l@{}}\textless{}event type\textgreater{}. \textless{}main argument type\textgreater{}, \\ \textless{}main argument span\textgreater{}. \textless{}sub-argument type\textgreater{}?\end{tabular} & \textbf{76.92} & 78.40 & 77.00 & 84.22 \\\midrule
\textless{}sub-argument query\textgreater{}? & 72.21 & 74.40 & 73.60 & 80.87 \\\midrule
\textless{}sub-argument query\textgreater$\:$in \textless{}event type\textgreater{}? & 72.03 & 75.14 & 74.26 & 81.68 \\\midrule
\textless{}sub-argument query\textgreater$\:$in \textless{}main argument span\textgreater{}? & 76.56 & 78.98 & 76.70 & 83.86 \\\midrule
\begin{tabular}[c]{@{}l@{}}\textless{}event type\textgreater{}. \textless{}main argument type\textgreater{},\\ \textless{}main argument span\textgreater{}. \textless{}sub-argument query\textgreater{}?\end{tabular} & 76.38 & \textbf{79.43} & \textbf{77.23} & \textbf{84.57} \\\bottomrule
\end{tabular}}
\caption{Sub-argument extraction (classification) results with different templates for extractive and generative QA methods. }
\label{tab:temp-sub}
\end{table*}


\section{Sampled Error Cases}
\label{apd:cases}
Table \ref{tab:apd-example} lists some example error cases as complementary material for the discussion in Section \ref{sec:results}. We present one example for each argument type in the table.

\clearpage
\onecolumn
\begin{xltabular}{\textwidth}{p{0.6\textwidth}|p{0.38\textwidth}}
\toprule
Input & Output \\ \midrule
\textbf{\underline{Query argument type:}} Subject \newline \textbf{\underline{Sentence:}} We report \textbf{a patient with inoperable pancreatic cancer} who developed gastrointestinal bleeding secondary to radiation-recall related to gemcitabine and review literature.& \textbf{\underline{Sequence Labelling:}} (ADE) {\color{red}a patient with}\newline  \textbf{\underline{Extractive QA:}} (ADE) a patient with inoperable pancreatic cancer\newline  \textbf{\underline{Generative QA:}} (ADE) a patient with inoperable pancreatic cancer \\\midrule
\textbf{\underline{Query argument type:}} Treatment\newline \textbf{\underline{Sentence:}} Supravenous hyperpigmentation in association with \textbf{CHOP chemotherapy} of a CD30 (Ki-1)-positive anaplastic large-cell lymphoma. & \textbf{\underline{Sequence Labelling:}} (ADE) {\color{red}CHOP chemotherapy of a CD30 (}\newline  \textbf{\underline{Extractive QA:}} (ADE) CHOP chemotherapy\newline  \textbf{\underline{Generative QA:}} (ADE) CHOP chemotherapy \\\midrule
\textbf{\underline{Query argument type:}} Effect \newline \textbf{\underline{Sentence:}} CONCLUSIONS: \textbf{Priapism} is an uncommon but potentially serious adverse effect of zuclopenthixol that practitioners, as with many other antipsychotics, should be aware of.& \textbf{\underline{Sequence Labelling:}} (ADE) Priapism\newline  \textbf{\underline{Extractive QA:}} ({\color{red}None}) {\color{red}None}\newline  \textbf{\underline{Generative QA:}} (ADE) Priapism \\\midrule
 \textbf{\underline{Query argument type:}} Subject.Age \newline \textbf{\underline{Sentence:}} CONCLUSIONS: Musculoskeletal complaints were the presenting symptoms in four of 44 \textbf{children} (9\%) treated for relapsed Wilms' tumors with ifosfamide, a derivative of cyclophosphamide.& \textbf{\underline{Sequence Labelling:}} ({\color{red}None}) children\newline  \textbf{\underline{Extractive QA:}} ({\color{red}None}) {\color{red}None}\newline  \textbf{\underline{Generative QA:}} (ADE) children \\\midrule
 \textbf{\underline{Query argument type:}} Subject.Gender\newline \textbf{\underline{Sentence:}} We report the case of a 74-year-old \textbf{female} patient who received the antide-pressant amitriptyline because of a major depression.& \textbf{\underline{Sequence Labelling:}} ({\color{red}ADE}) female\newline  \textbf{\underline{Extractive QA:}} ({\color{red}None}) {\color{red}None}\newline  \textbf{\underline{Generative QA:}} ({\color{red}ADE}) female \\\midrule
\textbf{\underline{Query argument type:}} Subject.Population \newline \textbf{\underline{Sentence:}}  Musculoskeletal complaints were the presenting symptoms in \textbf{four of 44} children (9\%) treated for relapsed Wilms' tumors with ifosfamide, a derivative of cyclophosphamide.& \textbf{\underline{Sequence Labelling:}} ({\color{red}None}) four of 44; {\color{red}(9\%}\newline  \textbf{\underline{Extractive QA:}} ({\color{red}None}) {\color{red}None}\newline  \textbf{\underline{Generative QA:}} (ADE) four of 44 \\\midrule
\textbf{\underline{Query argument type:}} Subject.Race \newline \textbf{\underline{Sentence:}}  CASE SUMMARY: A febrile 36-year-old seaman from \textbf{Mumbai (Bombay)} was prescribed \textgreater{}5 times the usual dose of chloroquine for malaria diagnosed empirically onboard ship. & \textbf{\underline{Sequence Labelling:}} (PTE) {\color{red}None}\newline  \textbf{\underline{Extractive QA:}} (PTE) {\color{red}None}\newline  \textbf{\underline{Generative QA:}} (PTE) {\color{red}None} \\\midrule
 \textbf{\underline{Query argument type:}} Subject.Disorder\newline \textbf{\underline{Sentence:}} We describe a 57-year-old man with \textbf{acral erythrocyanosis} progressing to acute digital ischemia and gangrene that developed after combined chemotherapy (bleomycin and methotrexate) used to treat a metastatic squamous cell carcinoma of the hypopharynx.& \textbf{\underline{Sequence Labelling:}} (ADE) {\color{red}None}\newline \textbf{\underline{Extractive QA:}} (ADE) {\color{red}None}\newline  \textbf{\underline{Generative QA:}} (ADE) {\color{red}None} \\\midrule
 \textbf{\underline{Query argument type:}} Treatment.Drug\newline\textbf{\underline{Sentence:}}  We conclude that \textbf{MB} is an effective treatment for \textbf{ifosfamide}-induced encephalopathy.\newline \textbf{\underline{Note:}} Nested events are included in this case. \newline MB is the Treatment.Drug for the PTE event, ifosfamide is the Treatment.Drug for the ADE event. & \textbf{\underline{Sequence Labelling:}} {\color{red}(ADE) ifosfamide}\newline  \textbf{\underline{Extractive QA:}} {\color{red}(ADE) ifosfamide}\newline  \textbf{\underline{Generative QA:}} {\color{red}(PTE) MB} \\\midrule
 \textbf{\underline{Query argument type:}} Treatment.Dosage\newline \textbf{\underline{Sentence:}} Severe rhabdomyolysis following \textbf{massive} ingestion of oolong tea: caffeine intoxication with coexisting hyponatremia.& \textbf{\underline{Sequence Labelling:}} (ADE) {\color{red}None}\newline  \textbf{\underline{Extractive QA:}} (ADE) {\color{red}None}\newline  \textbf{\underline{Generative QA:}} (ADE) {\color{red}None} \\\midrule
 \textbf{\underline{Query argument type:}} Treatment.Route \newline \textbf{\underline{Sentence:}} Three hundred and thirty eight patients with moderate to severe painful diabetic neuropathy despite receiving their maximum tolerated dose of gabapentin, had \textbf{oral} prolonged-release oxycodone or placebo \textbf{tablets} added to their therapy for up to 12 weeks. & \textbf{\underline{Sequence Labelling:}} ({\color{red}None}) oral; tablets\newline  \textbf{\underline{Extractive QA:}} ({\color{red}None}) {\color{red}None}\newline  \textbf{\underline{Generative QA:}} (PTE) {\color{red}None} \\\midrule
 \textbf{\underline{Query argument type:}} Treatment.Frequency \newline\textbf{\underline{Sentence:}}  A 36-y-o patient with schizophrenia, who had consumed gradually increasing quantities of oolong tea that eventually reached 15 L \textbf{each day}, became delirious and was admitted to a psychiatric hospital.& \textbf{\underline{Sequence Labelling:}} (ADE) each day\newline  \textbf{\underline{Extractive QA:}} (ADE) {\color{red}None}\newline  \textbf{\underline{Generative QA:}} (ADE) {\color{red}None} \\\midrule
\textbf{\underline{Query argument type:}} Treatment.Duration \newline  \textbf{\underline{Sentence:}} A 10-year-old boy with osteosarcoma and normal renal function manifested laboratory evidence of impending renal toxicity and extreme elevation of aspartate aminotrasferase and alanine aminotransferase within 2 hours after the completion of a \textbf{4-hour} infusion of high-dose methotrexate (MTX) (12 g/m2), and went on to develop acute renal failure with life-threatening hyperkalemia 29 hours later. & \textbf{\underline{Sequence Labelling:}} (ADE) {\color{red}hour}\newline  \textbf{\underline{Extractive QA:}} (ADE) {\color{red}None}\newline  \textbf{\underline{Generative QA:}} (ADE) {\color{red}None} \\\midrule
 \textbf{\underline{Query argument type:}} Treatment.Time\_elapsed \newline \textbf{\underline{Sentence:}}  She was placed on adjuvant Adriamycin (doxorubicin) chemotherapy, but \textbf{6 months} later died of Adriamycin toxicity.& \textbf{\underline{Sequence Labelling:}} ({\color{red}None}) 6 month {\color{red}later}\newline  \textbf{\underline{Extractive QA:}} ({\color{red}None}) {\color{red}None}\newline  \textbf{\underline{Generative QA:}} (ADE) {\color{red}None} \\\midrule
\textbf{\underline{Query argument type:}} Treatment.Disorder \newline \textbf{\underline{Sentence:}}  Pulmonary edema during acute infusion of epoprostenol in a patient with \textbf{pulmonary hypertension and limited scleroderma}.& \textbf{\underline{Sequence Labelling:}} (ADE) pulmonary hypertension and limited scleroderma\newline  \textbf{\underline{Extractive QA:}} (ADE) {\color{red}epoprostenol}\newline  \textbf{\underline{Generative QA:}} (ADE) pulmonary hypertension; limited scleroderma \\\midrule
 \textbf{\underline{Query argument type:}} Combination.Drug \newline \textbf{\underline{Sentence:}}  Thus, tardive seizures in our cases are thought to be related to \textbf{piperacillin} and \textbf{cefotiam}.& \textbf{\underline{Sequence Labelling:}} (ADE) piperacillin; cefotiam \newline  \textbf{\underline{Extractive QA:}} (ADE) {\color{red}piperacillin}\newline  \textbf{\underline{Generative QA:}} (ADE) piperacillin; cefotiam \\ \bottomrule
\caption{Example error cases. Standard gold annotations are bolded in the original sentence. Model detected event types are shown in ($\cdot$) before the predicted argument span. Error predictions are shown in red.}
\label{tab:apd-example}
\end{xltabular}
\clearpage
\twocolumn

\end{document}